\documentclass{article}
\usepackage[preprint]{neurips_2025}

\usepackage{natbib}
\usepackage{graphicx}
\usepackage[utf8]{inputenc} 
\usepackage[T1]{fontenc}    
\usepackage{hyperref}       
\usepackage{url}            
\usepackage{booktabs}       
\usepackage{algorithm}
\usepackage{mypython}
\usepackage{amsfonts}       
\usepackage{amsmath}
\usepackage{amsthm}
\usepackage{amssymb}
\usepackage{nicefrac}       
\usepackage{microtype}      
\usepackage{xcolor}         
\usepackage[font=small,labelfont=bf]{caption}
\usepackage{enumitem}
\usepackage{wrapfig}
\usepackage{listings}
\usepackage{caption}
\usepackage{multirow}
\usepackage[most]{tcolorbox}
\usepackage[normalem]{ulem}
\usepackage{xspace}
\usepackage{float}
\usepackage{tabularx}
\usepackage{minted}
\usepackage{booktabs}
\usepackage[normalem]{ulem}
\usepackage{xcolor} 
\usepackage{soul}   
\usepackage{svg}
\usepackage{array}
\usepackage{longtable}
\usepackage{adjustbox}
\usepackage{makecell}
\usepackage{pifont}
\usepackage{threeparttable}
\newcommand{\xmark}{\ding{55}}%

\definecolor{codegreen}{rgb}{0,0.6,0}
\definecolor{codegray}{rgb}{0.5,0.5,0.5}
\definecolor{codepurple}{rgb}{0.58,0,0.82}
\definecolor{backcolour}{rgb}{0.95,0.95,0.95}

\lstdefinestyle{mystyle}{
    backgroundcolor=\color{backcolour},   
    commentstyle=\color{codegreen},
    keywordstyle=\color{magenta},
    numberstyle=\tiny\color{codegray},
    stringstyle=\color{codepurple},
    basicstyle=\ttfamily\footnotesize,
    breakatwhitespace=false,         
    breaklines=true,                 
    captionpos=b,                    
    keepspaces=true,                 
    numbers=left,                    
    numbersep=5pt,                  
    showspaces=false,                
    showstringspaces=false,
    showtabs=false,                  
    tabsize=2
}

\useunder{\uline}{\ul}{}

\newcommand{\modelname}{Ministral~3\xspace}
\newcommand{\methodname}{Cascade Distillation\xspace}

\newif\ifhidecomment
\hidecommentfalse

\definecolor{mygreybg}{gray}{0.95}

\sethlcolor{mygreybg}

\title{\modelname}

\begin{document}

\maketitle

\vspace{-0.1in}
\begin{center}
\vspace{-45pt}
\centering
\includegraphics[width=0.8\linewidth,keepaspectratio]{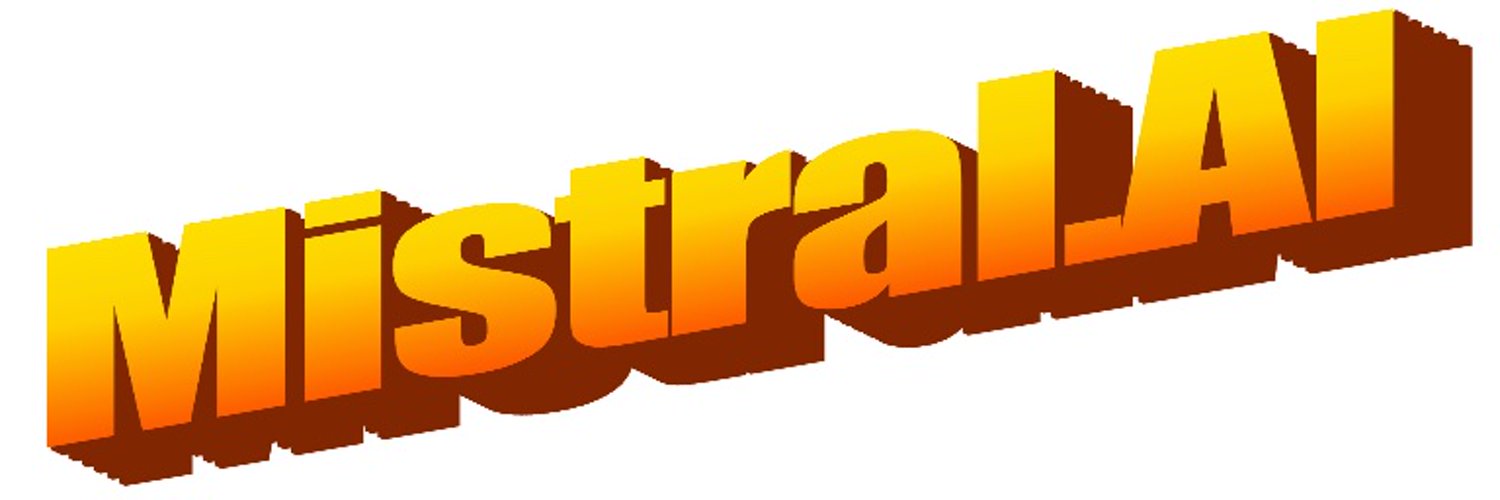}
\end{center}

\begin{abstract}
We introduce the \modelname series, a family of parameter-efficient dense language models designed for compute and memory constrained applications, available in three model sizes: 3B, 8B, and 14B parameters.
For each model size, we release three variants: a pretrained base model for general-purpose use, an instruction finetuned, and a reasoning model for complex problem-solving.
In addition, we present our recipe to derive the \modelname~models through \methodname, an iterative pruning and continued training with distillation technique.
Each model comes with image understanding capabilities, all under the Apache 2.0 license.

\textbf{Webpage:} {\small \url{https://mistral.ai/news/mistral-3}} \\
\textbf{Models:} {\small \url{https://huggingface.co/collections/mistralai/ministral-3}} \\
\end{abstract}
\vspace{0.5cm}

\section{Introduction}

\looseness=-1 In this work, we introduce \modelname, a family of dense models trained in a compute- and data-efficient manner through iterative shrinking and distillation from a parent pretrained model.
Unlike popular pretrained models such as Qwen3 \citep{yang2025qwen3technicalreport} or Llama3 \citep{dubey2024llama} that are trained on 36 trillion and 15 trillion tokens respectively, we are able to produce competitive models trained for between 1 and 3 trillion tokens by leveraging Mistral Small 3.1, a strong 24B-parameter parent model.

Available in three sizes: 3B, 8B, and 14B parameters, all \modelname~models are descendants of Mistral Small 3.1\footnote{\url{https://mistral.ai/news/mistral-small-3-1}}, obtained via a \methodname approach. We present three variants for each model size: base, instruct, and reasoning, each with image understanding capabilities and context lengths up to 256k tokens (128k for reasoning models).

A key component of \modelname is our \methodname training strategy, an iterative pruning and distillation method, which progressively transfers pretrained knowledge from a large parent model down to a family of compact children models. Our recipe allows us to achieve performance that is competitive with models which had a much larger training budget. For example, the \modelname~14B Base model closely matches Mistral Small 3.1 Base, while being more than 40\% smaller and trained on a much shorter horizon.

\looseness=-1 After post-training, we achieve competitive results with similarly sized open weight models such as Gemma~3 ~\citep{kamath2025gemma3technicalreport}, Qwen~3~\citep{yang2025qwen3technicalreport,bai2025qwen3vltechnicalreport}, and Mistral~Small~3.2~2506.

\begin{figure}[t]
\includegraphics[width=\linewidth,keepaspectratio]{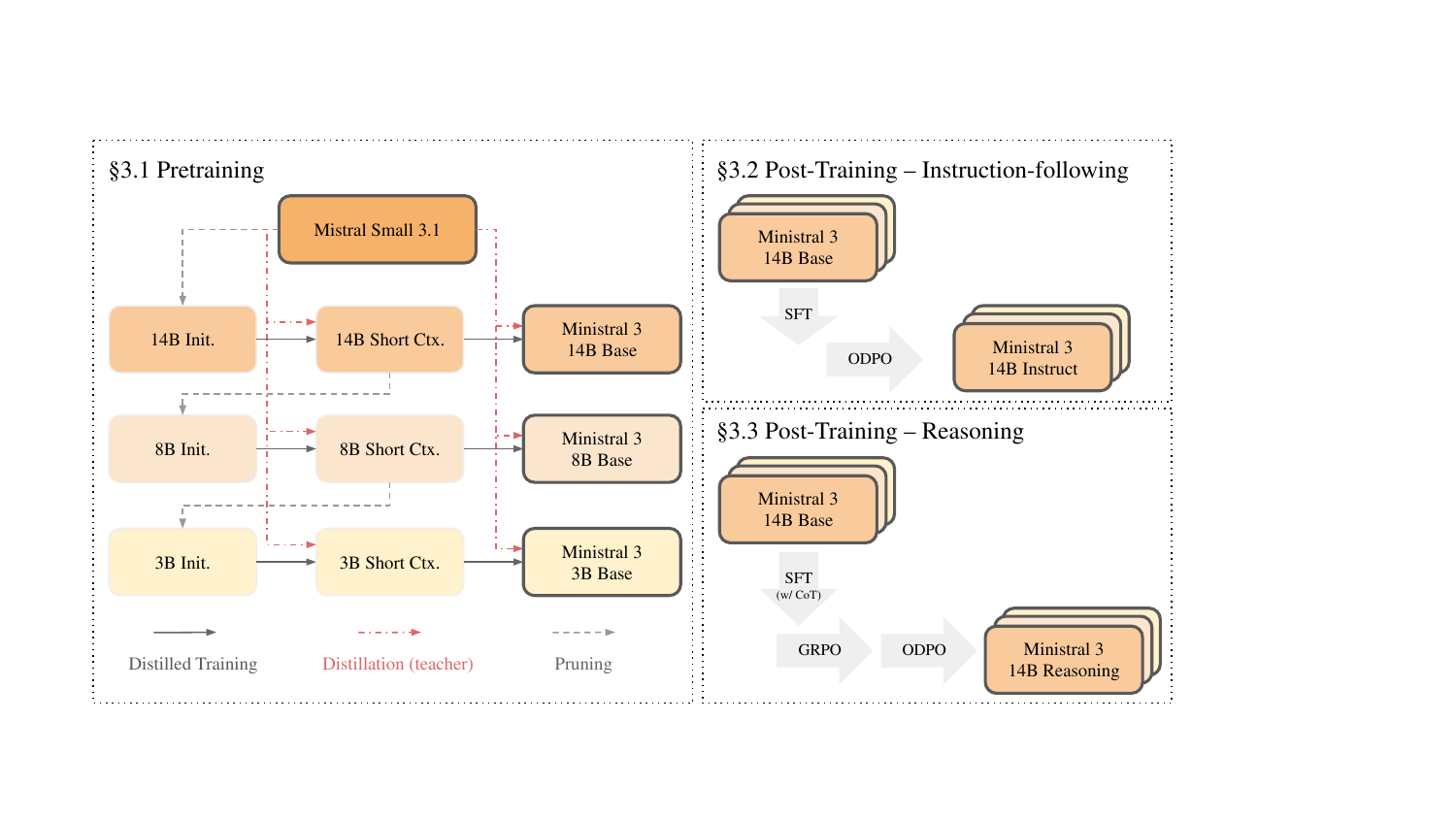}
\caption{
\small
\looseness=-1 \textbf{Overview of \modelname~training recipe.}
\textbf{Pretraining}: We start from pruning the parent model, Mistral Small 3.1, into the largest child model (14B Init.).
Next, we continue pretraining the child model with logit distillation from the parent model as the teacher to obtain the up-trained short context child model (14B Short Ctx.).
From 14B Short Ctx., we perform another round of distillation with longer context window (see \S\ref{subsec:pt} for details) to obtain the final \modelname~14B Base model.
In parallel, 14B Short Ctx. is pruned to initialize the next child model (8B Init.), from which we repeat the process to derive \modelname~8B Base model.
We repeat the same process for the 3B version.
\textbf{Post-training}:
Each Base model is then post-trained into the instruction-following and reasoning variants.
For instruction-following, our post-training recipe includes supervised fine-tuning (SFT) and Online Direct Preference Optimization (ODPO).
For reasoning, the process involved supervised fine-tuning with chain-of-thought data (SFT w/ CoT), Group Relative Policy Optimization (GRPO;~\citet{shao2024deepseekmathpushinglimitsmathematical}), and ODPO.
}
\label{fig:method}
\end{figure}

The main contributions can be summarized as follows:
\begin{itemize}
\item We introduce \modelname, a family of 9 dense language models - a pretrained, an instruction finetuned, and a reasoning model, each at the 14B, 8B, and 3B parameter scales. All \modelname~models (3 sizes × 3 variants) are open-weight under the Apache 2.0 license.
\item We present a compute-efficient pretraining recipe, \methodname, with which these models have been pretrained at a fraction of the cost it would take to pretrain from scratch.
\item We independently confirm findings from prior work that (a) there exists a "capacity gap" where a stronger teacher does not yield a stronger student model for pretraining, but post-training continues to benefit from a stronger teachers (b) distilling from a post-trained as opposed to a pretrained teacher when pretraining the student model results in better benchmark scores (c) distilling from a human preference optimized teacher is better than one that has only been post-trained with SFT.
\end{itemize}

\section{Model Architecture}
\label{sec:method}
\vspace{-10pt}
\begin{table}[h]
\centering
\begin{threeparttable}
\caption{
\small
\label{tab:model_specs}
Architectural specifications and hyperparameters for the \modelname family. All models use a vocabulary size of 131K tokens.
}
\small
\begin{tabular}{@{}lcccccc@{}}
\toprule
  & \multirow{2}{*}{\textbf{Layers}} & \textbf{Latent} & \textbf{Q/KV} & \textbf{FFN} &  \textbf{Tied} & \textbf{Context} \\
  & & \textbf{dim.} & \textbf{heads} & \textbf{dim.} &  \textbf{Embeddings} & \textbf{Length} \\
\midrule
\modelname 14B  & 40 & 5120  & 32~/~8 & 16384  & \xmark & 256k \\
\modelname 8B  & 34 & 4096  & 32~/~8 & 14336  &  \xmark & 256k \\
\modelname 3B  & 26 & 3072  & 32~/~8 & 9216   &  \checkmark & 256k \\
\bottomrule
\end{tabular}
\end{threeparttable}
\end{table}

The \modelname family is based on the decoder-only transformer architecture~\citep{vaswani2017attention}.
All models share a common architectural foundation with size-specific scaling.
As shown in Table~\ref{tab:model_specs}, the family consists of three sizes: 3B, 8B, and 14B parameters, with 26, 34, and 40 layers respectively.
Other architectural choices include Grouped Query Attention ~\citep{ainslie2023gqa} with 32 query heads and 8 key-value heads, RoPE~\citep{su2021roformer} positional embeddings, SwiGLU activation~\citep{shazeer2020glu}, and RMSNorm~\citep{zhang2019root}. For long-context extension, we use YaRN ~\citep{peng2023yarn} and position-based softmax temperature scaling in the attention layer ~\citep{nakanishi2025scalable, meta2025llama4}. The 3B model uses tied input-output embeddings to avoid embedding parameters dominating the overall parameter count. All models use a vocabulary of 131K tokens and support context lengths up to 256K tokens.

\textbf{Vision encoder.}
All \modelname models use a 410M parameter ViT as a vision encoder for image understanding that is copied from Mistral Small 3.1 Base and kept frozen, with the same architecture described in Pixtral~\citep{agrawal2024pixtral}. We discard the pretrained projection layer from the ViT to language model's space and train a new projection for every model.

\section{Training Recipe}

Figure~\ref{fig:method} illustrates the training pipeline of the \modelname~models, consisting of a pretraining followed by two distinct post-training phases to produce instruction finetuned and reasoning variants.

\subsection{Pretraining}
\label{subsec:pt}

\begin{figure}[!h]
\centering
\begin{minipage}{0.45\textwidth}
\begin{algorithm}[H]
   \caption{\methodname.}
   \label{alg:method}
\begin{python}[basicstyle=\ttfamily\bfseries]
model = MS3 # Mistral Small 3.1

for model_size in [14B, 8B, 3B]:

  # pruning (see Algo. 2)
  model = prune(model, model_size)

  # short context distillation
  model = model.train(
    data=short_data,
    teacher_model=MS3,
  )

  # long context distillation 
  final_model = model.train(
    data=long_data,
    teacher_model=MS3,
  )
  yield (model_size, final_model)
\end{python}
\end{algorithm}
\end{minipage}%
\hfill
\begin{minipage}{0.53\textwidth}
\centering
\includegraphics[width=\linewidth,keepaspectratio]{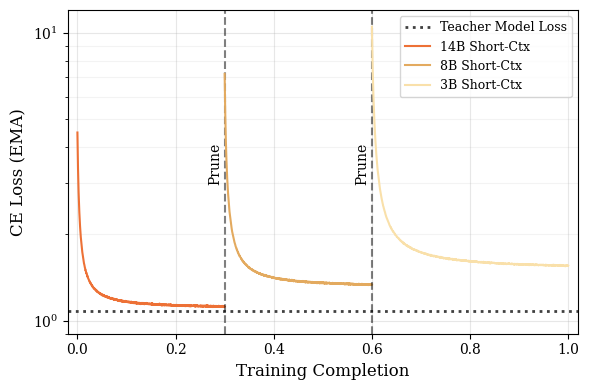}
\caption{Illustration of \methodname.}
\label{fig:distillation_cascade}
\end{minipage}
\end{figure}

\textbf{\methodname.}
Pretraining of the \modelname~models starts from the Mistral Small 3.1 Base (MS3.1) model.
We use \textit{\methodname}, an iterative approach to prune and distill MS3.1 into the smaller successors. 
\methodname is a compute-efficient process for pretraining children models of decreasing target sizes, given a pre-trained larger parent model.
As summarized in Algorithm~\ref{alg:method}, it relies on an iterative ``prune-distill-repeat'' approach:
\begin{enumerate}
    \item Prune: initialize the weights of a child model via pruning a larger pre-trained model.
    \item Distill: up-train the freshly pruned model via distillation from the teacher model's logits.
    \item Repeat: apply this strategy repeatedly to shrink the child model into something even smaller.
\end{enumerate}

Model pruning at each stage follows a similar approach to Minitron and Wanda ~\citep{sun2023simple,sreenivas2024minitron,muralidharan2024compact} with the distillation teacher being Mistral Small 3.1 for all variants. Details of pruning and distillation are provided in the following paragraphs.

Compared to training each small model from scratch, \methodname produces a model that is significantly more FLOP efficient.
It is also worth noting that the end-to-end process can be viewed as a form of continual pretraining of the parent model with weight pruning.
As illustrated in Figure~\ref{fig:distillation_cascade}, data repetition is avoided throughout the process as \methodname goes through the data mix in a single run with pruning en route.

\vspace{3pt}
\textbf{Pruning.}
Similar to Minitron, our pruning strategies are designed to preserve the most critical components of the original model (over a validation dataset) while reducing its size.
We employ following key pruning techniques:

\begin{itemize}[leftmargin=20pt]
\item \textbf{Layer Pruning:} Unlike \citet{sreenivas2024minitron}, which relies on counterfactual downstream perplexities from removing individual layers, we find that the ratio of input to output activation norms provides a simpler yet strong proxy for layer importance.

\item \textbf{Hidden Dimension Pruning:} Apply Principal Component Analysis (PCA) to concatenated activations from attention normalization and feed-forward normalization layers across all layers. This yields a single rotation matrix consistent across the entire network that projects the model to a lower-dimensional space while maximizing explained variance.

\item \looseness=-1 \textbf{Feedforward Dimension Pruning:} 
For MLPs with gated-linear activation functions such as SwiGLU~\citep{shazeer2020glu}, expressed as $W_2(SiLU(W_1x) * W_3x)$ given a very large batch $x$, we prune dimension of the matrices $W_1, W_2, W_3$.
To determine the columns of $W_1, W_3$ to keep, we compute the importance score defined as the averaged absolute value of each dim of the expression above.
We then keep only the corresponding rows of $W_2$ with the indices yielded above.

\end{itemize}

Algorithm~\ref{alg:pruning} provides more detail on our pruning strategy:
\begin{algorithm}[!h]
   \caption{Pruning stage of \methodname. It takes as input a pre-trained model and target size configuration to prune to. We use \texttt{input\_x} and \texttt{output\_x} to refer to activations from a large calibration batch.}
   \label{alg:pruning}

\begin{python}[basicstyle=\ttfamily\bfseries]
def prune(model, target_size):

    target_n_layers, target_dim, target_ffn_dim = get_config(target_size)

    # layer pruning
    scores = []
    for layer in model.layers:
        input_norm = layer.input_x.norm(dim=-1)
        output_norm = layer.output_x.norm(dim=-1)
        scores.append(
            (output_norm / input_norm).mean()
        )

    layers_to_keep = topk(scores, k=target_n_layers)
    model = remove_layers(model, layers_to_keep)

    # hidden dimension pruning
    norm_inputs = []
    for layer in model.layers:
        norm_inputs.extend([
            layer.attn_norm.input_x,
            layer.ffn_norm.input_x,
        ])
    
    rotation = PCA(norm_inputs, n_components=n_dims)
    model = apply_rotation(model, rotation, target_dim)

    # feedforward pruning
    for layer in model.layers:
        importance = abs(
            silu(layer.ffn.w1.output_x) * layer.ffn.w3.output_x
        ).mean(dim=(0,1))
        dims_to_keep = topk(importance, k=target_ffn_dim)
        layer.ffn = prune_hidden_dims(layer.ffn, dims_to_keep)


  return model
\end{python}

\end{algorithm}

\vspace{3pt}
\textbf{Distillation.}
After weight initialization, each child model is trained on a mixture of text-only and interleaved text with image data with logit distillation from a teacher model. We find that training with just the forward KL distillation objective outperforms tuning the coefficients of an objective that weights the distillation objective and the next token prediction objective differently.  For all stages and model sizes, we use the parent model as the teacher model(more details in \S\ref{discuss:teacher}).

The pretraining phase consists of a two-stages:

\begin{enumerate}[leftmargin=20pt,label=(\arabic*)]
    \item \textbf{Short context stage} with a context window of length 16,384. The output of this phase is the input to to the pruning phase of the next child model.
    \item \textbf{Long context stage} to extend the context window from 16,384 to 262,144 using YaRN~\citep{peng2023yarn} and position-based temperature scaling~\citep{nakanishi2025scalable,meta2025llama4}.
\end{enumerate}

\subsection{Post-Training: Ministral Instruct}

\looseness=-1 To impart instruction-following capabilities \citep{ouyang2022training}, pretrained models are fine-tuned using a curated dataset comprising high-quality multimodal and text-only instruction-following data. The fine-tuning phase also consists of two stages: Supervised Fine-Tuning (SFT) and Online Direct Preference Optimization (ODPO).

\vspace{3pt}
\subsubsection{Supervised Fine-tuning}
\looseness=-1 We run SFT with fp8 quantization, using a logit distillation loss from a strong teacher. Unlike pretraining, each model is distilled from Mistral Medium 3 model (more details in \S\ref{discuss:teacher}). Similar to the pretraining phase, the vision encoder remains frozen while the adapter is trainable.

\vspace{3pt}
\subsubsection{Online Direct Preference Optimization stage}
Direct Preference Optimization (DPO) \citep{rafailov2023dpo} offers a lightweight framework for human preference optimization by learning directly from offline pairwise preferences. For the \modelname~models, we adopt its online variant, Online Direct Preference Optimization (ODPO) \citep{guo2024directlanguagemodelalignment} where, for each example, we sample two candidate responses from the current policy with temperature $T{=}0.7$, and use a text-based reward model to rank the responses.

This method relies on a Pairwise Reward Model (PWRM) to dynamically rank candidate responses. The PWRM is trained via supervised fine-tuning (SFT) on structured pairwise data: given a conversation history and two candidate responses, it predicts which response is preferred. In addition, we refine the classic DPO loss by incorporating the binomial probabilistic output of the PWRM, replacing hard winner/loser labels with a two-sided loss that weights each response by its probability of being preferred. We make two additional changes to stabilize the learning process: (1) we adjust the PWRM temperature to calibrate the win / loss probabilities; and (2) we employ a $\beta$-rescaling technique, allowing for a more beta-invariant rescaling of dpo loss.

In practice, the online variant is particularly important for mitigating model-induced artifacts, such as infinite generations. This is also facilitated by some heuristics, such as automatically treating any response that exhibits an infinite loop during sample as “loser,” preventing such behavior from being reinforced. Finally, we enable tool execution during generation, which improves the model’s tool-use performance.

In summary, we found that using online preference optimization improves alignment with human preferences significantly over both the SFT and offline variants. We release the models resulting from this phase as \modelname-14B/8B/3B Instruct.

\subsection{Post-Training: Ministral Reasoning}

Post-training for reasoning models begins from the pre-trained checkpoint as opposed to the ODPO variant. We train the model for inference-time scaling using a three-stage pipeline composed of SFT, GRPO and ODPO, using the long-context pretrained checkpoint as the starting point. Models released after this reasoning-oriented fine-tuning stage are referred to as \modelname 14B/8B/3B Reasoning.

\subsubsection{Reasoning Supervised Fine-Tuning}

In this stage, the model is finetuned on a mixture of short and long CoT samples. The former is derived from our general SFT data mixture whereas the latter consists of reasoning traces which have been prefixed with a reasoning specific system prompt. 

\looseness=-1 The reasoning traces come from a diverse set of domains including mathematics, coding, general dialogue, instruction following, multilingual tasks, tool use, and visual reasoning. We apply lightweight filtering to remove examples that are poorly formatted, contain excessive repetition, or have undesirable language switching, ensuring that the model is exposed to clean and well-structured chains of thought.

\textbf{3B SFT}: For the 3B model, vanilla SFT led to a brittle, overly verbose model with lots of repetition and infinite generations in its output. To mitigate this, we did logit distillation with Magistral Small 1.2 as teacher. This helped reduce verbosity and stabilized subsequent RL training.

\subsubsection{Reinforcement Learning}

\looseness=-1 We perform GRPO~\citep{deepseekai2025deepseekr1} on top of the SFT checkpoint to refine the model's thinking and improve the performance further on reasoning tasks. The training is conducted in two stages:

\textbf{STEM RL}: In the first stage, we train the model on math, code and visual-reasoning tasks. We collect question-answer pairs from a diverse set of open and proprietary sources. The samples are filtered and cleaned using a rigorous multi-step pipeline (detailed in \cite{rastogi2025magistral}) to remove invalid, incomplete and very easy/hard problems.

\looseness=-1 \textbf{General RL}: In the second stage, we broaden the scope beyond STEM problems. We generate atomic grading rubrics for a diverse set of prompts including general chat, instruction-following, and open-ended reasoning tasks. During GRPO, an LLM judge evaluates each model rollout against these rubrics (e.g., faithfulness to the prompt, response quality) and the final reward is set to the fraction of satisfied heuristics. This stage improves the instruction following and general chat capabilities of the model while maintaining, and sometimes even improving, the performance on the STEM benchmarks.

For both stages, we follow the GRPO training recipe from \cite{rastogi2025magistral}.The maximum generation length is increased from 32K to 80K, since we observed a non-trivial proportion of truncated generations during RL. Allowing longer outputs allowed the model to finish its reasoning for the most challenging problems, resulting in additional performance gains.

\subsubsection{Online Direct Preference Optimization}

Finally, we apply ODPO as a post-RL alignment stage to better align with user preferences and polish the model’s conversational and instructional behavior. The overall procedure follows the same setup as used for our non-reasoning instruct models, with one modification – The thinking chunks are stripped from the model's generations before sending them to the reward model for scoring. Some additional experimental details are discussed in Section \ref{discuss:odpo}.

\section{Results}

In this section, we report the results of \modelname~models on a variety of benchmarks.
We also compare \modelname to other open-weight models on the same scale, namely the Qwen 3 family~\citep{yang2025qwen3technicalreport,bai2025qwen3vltechnicalreport} and the Gemma 3 family~\citep{kamath2025gemma3technicalreport}.
For external models, we re-run all benchmarks with our own evaluation pipeline for fair comparison.

\looseness=-1 We evaluated on the following benchmarks:
\textbf{General:} MMLU~\citep{hendrycks2020measuring}, MMLU-Redux~\citep{perez2024mmluRedux}, ARC-Challenge~\citep{clark2018think}, RACE High~\citep{lai2017race}, TriviaQA~\citep{joshi2017triviaqa}, NaturalQS~\citep{kwiatkowski2019natural},
and AGIEval~\citep{zhong2023agieval}.
\textbf{Math \& Code:} MATH~\citep{hendrycks2021measuring}, GPQA Diamond~\citep{rein2024gpqa}, and MBPP~\citep{austin2021program}.
\textbf{Multimodal:} MMMU~\citep{yue2023mmmu} and MathVista~\citep{lu2024mathvista}.
\textbf{Post-training:} Arena Hard~\citep{li2024arenahard}, WildBench~\citep{lin2024wildbench}, MM MTBench\footnote{https://huggingface.co/datasets/mistralai/MM-MT-Bench}, AIME 2024/2025, HMMT 2025, PhyBench~\citep{liu2025phybench}, and LiveCodeBench~\citep{jain2024livecodebench}.

\begin{table*}[t!]
\centering
\caption{
\label{tab:pt_compare}
Comparing \modelname~Base models against the Gemma 3 base models and the Qwen 3 base models on pretraining benchmarks. All the results are reported after running the evaluations using our internal harness with identical configuration. 
}
\resizebox{\textwidth}{!}{
\begin{tabular}{lccccc}
\toprule
\multirow{2}{*}{Model} & {MMLU-Redux} & {TriviaQA} & {MATH} & {AGIEval} & {Multilingual MMLU} \\
& {\scriptsize (5-shot)} & {\scriptsize (5-shot)} & {\scriptsize (CoT 2-Shot)} & {\scriptsize (5-shot)} & {\scriptsize (5-Shot)} \\
\midrule
\midrule
Qwen 3 14B & 83.7 & 70.3 & 62.0 & 66.1 & 75.4 \\
\modelname 14B & 82.0 & 74.9 & 67.6 & 64.8 & 74.2 \\
\midrule
Gemma 3 12B & 76.6 & 78.8 & 48.7 & 58.7 & 69.0 \\
Qwen 3 8B & 79.4 & 63.9 & 57.6 & 59.6 & 70.0 \\
\modelname 8B & 79.3 & 68.1 & 62.6 & 59.1 & 70.6 \\
\midrule
Gemma 3 4B & 62.6 & 64.0 & 29.4 & 43.0 & 51.6 \\
Qwen 3 4B & 75.9 & 53.0 & 40.5 & 57.0 & 67.7 \\
\modelname 3B & 73.5 & 59.2 & 60.1 & 51.1 & 65.2 \\
\midrule
\bottomrule
\end{tabular}
}
\end{table*}

\begin{table*}[t!]
\centering
\resizebox{\textwidth}{!}{
\begin{threeparttable}
\caption{
\label{tab:pt_full}
Evaluation results of the \modelname~Base family compared to the teacher model Mistral Small 3.1 24B across general reasoning, math \& code, multilingual, and multimodal benchmarks. Performance scales smoothly with model size, yet the pruned \modelname~variants retain a large fraction of the teacher’s capability despite substantial parameter reductions.
}
\begin{tabular}{lcccc}
\toprule
Evaluation & Mistral Small 24B & \modelname 14B & \modelname 8B & \modelname 3B \\
\midrule
\midrule
\textit{General} & & & & \\
\midrule
~~MMLU {\scriptsize (5-shot)} & 81.0 & 79.4 & 76.1 & 70.7 \\
~~MMLU-Redux {\scriptsize (5-shot)} & 82.7 & 82.0 & 79.3 & 73.5 \\
~~ARC-Challenge & 91.6 & 89.9 & 88.0 & 85.5 \\
~~RACE High & 52.1 & 52.3 & 49.7 & 49.3 \\
~~TriviaQA {\scriptsize (5-shot)} & 79.3 & 74.9 & 68.1 & 59.2 \\
~~NaturalQS {\scriptsize (5-shot)} & 34.4 & 29.9 & 25.8 & 21.9 \\
\midrule
\textit{Math \& Code}  & & & & \\
\midrule
~~MATH {\scriptsize (CoT 2-Shot)} & 55.8 & 67.6 & 62.6 & 60.1 \\
~~GPQA Diamond {\scriptsize (0-shot)} & 36.9 & 39.9 & 39.9 & 33.8 \\
~~MBPP {\scriptsize (3-shot Pass@1)} & 71.6 & 71.6 & 70.0 & 63.0 \\
\midrule
\textit{Multilingual MMLU}  & & & & \\
\midrule
~~European avg.$^\text{\textdagger}$ ~{\scriptsize (5-shot)} & 78.8 & 76.9 & 73.4 & 68.4 \\ 
~~Chinese {\scriptsize (5-shot)} & 75.7 & 75.1 & 71.3 & 64.1 \\
~~Japanese {\scriptsize (5-shot)} & 76.7 & 75.9 & 72.2 & 65.7 \\
~~Korean {\scriptsize (5-shot)} & 59.3 & 59.0 & 55.3 & 48.9 \\
\midrule
\textit{Multimodal} & & & & \\
\midrule
~~MMMU {\scriptsize (2-shot)} & 59.1 & 59.9 & 55.1 & 52.4 \\
~~MathVista & 51.3 & 43.6 & 35.7 & 23.3 \\
\midrule
\bottomrule
\end{tabular}
\begin{tablenotes}
    \item[\textdagger] Averaged over German, Spanish, French, Italian, and Portuguese.
\end{tablenotes}
\end{threeparttable}
}
\end{table*}

\subsection{Pretraining Results}

In Table~\ref{tab:pt_compare}, we compare \modelname~Base models against other open-weight models of similar size from the Gemma 3 family and the Qwen 3 family. 

At the 14B scale, \modelname~demonstrates strong performance, outperforming Qwen 3 14B on TriviaQA and MATH, while being competitive on other benchmarks.
Our 14B model is also significantly better than Gemma 12B across all benchmarks. At the 8B scale, we observe a similar trend. It is also worth pointing out that \modelname~8B outperforms the larger Gemma 12B in most of the evaluations (except TrivaiQA), highlighting the strong parameter efficiency of \modelname~ models.

At the 3B scale, the same overall trend persists, but performance gaps between models become more pronounced.
Additional pretraining evaluation results for \modelname~Base models along with the teacher model are provided in Table~\ref{tab:pt_full}.

\subsection{Post-training Results}

\begin{table*}[t!]
\centering
\caption{
\label{tab:ins}
Performance comparison of \modelname instruct models against instruction-tuned baselines from the Qwen 3 and Gemma 3 families. Models are grouped by size to facilitate like-for-like comparisons.
}
\begin{tabular}{lcccc}
\toprule
Model                     & Arena Hard & WildBench & MATH {\scriptsize (maj@1)} & MM MTBench \\
\midrule
Qwen3 14B (Non-Thinking)  & 42.7       & 65.1      & 87.00      & N/A \\
Ministral 3 14B           & 55.1       & 68.5      & 90.40      & 84.90      \\
\midrule
Gemma3-12B-Instruct       & 43.6       & 63.2      & 85.40      & 67.00      \\
Qwen3-VL-8B-Instruct      & 52.8       & 66.3      & 94.60      & 80.00      \\
Ministral 3 8B            & 50.9       & 66.8      & 87.60      & 80.80      \\
\midrule
Gemma3-4B-Instruct        & 31.8       & 49.1      & 75.90      & 52.30      \\
Qwen3-VL-4B-Instruct      & 43.8       & 56.8      & 90.00      & 80.08      \\
Ministral 3 3B            & 30.5       & 56.8      & 83.00      & 78.30      \\
\midrule
Qwen3-VL-2B-Instruct      & 16.3       & 42.2      & 78.60      & 63.60      \\
\bottomrule
\end{tabular}
\end{table*}

In Table~\ref{tab:ins}, we compare \modelname~Instruct models against Instruct models from the Gemma 3 family and the Qwen 3 family.
For Qwen 3, we report the results for the latest vision enabled instruct variants (Qwen3-VL).

In Table~\ref{tab:reasoning}, we compare \modelname~Reasoning models against reasoning models from the Qwen 3 family.
To ensure a fair comparison, all models are evaluated using the same evaluation pipeline. To reduce variance, we report pass@16 except LiveCodeBench which is evaluated using pass@5.

\begin{table*}[t!]
\centering
\caption{
\label{tab:reasoning}
Comparison of \modelname reasoning models with size-matched Qwen 3 reasoning counterparts on mathematics, science, and code benchmarks.
}
\resizebox{\textwidth}{!}{
\begin{tabular}{l|cc|cc|cc}
\toprule
\multirow{2}{*}{Benchmark}    & Qwen 3 & Ministral 3 & Qwen3-VL & Ministral 3  & Qwen3-VL & Ministral 3 \\
& 14B &  14B & 8B & 8B & 4B & 3B\\
\midrule
AIME 2024                  & 83.7                 & 89.8            & 86.0                 & 86.0           & 72.9                 & 77.5           \\
AIME 2025                 & 73.7                 & 85.0            & 79.8                 & 78.7           & 69.7                 & 72.1           \\
HMMT 2025                  & 55.8                 & 67.5            & 57.5                 & 55.8           & 50.8                 & 51.7           \\
GPQA Diamond            & 66.3                 & 71.2            & 67.1                 & 66.8           & 60.1                 & 53.4           \\
PhyBench                & 22.0                 & 26.0            & 22.0                 & 20.0           & 9.0                  & 15.0           \\
LiveCodeBench v6        & 59.3                 & 64.6            & 58.0                 & 61.6           & 51.3                 & 54.8           \\
\bottomrule
\end{tabular}
}
\end{table*}

\section{Discussions}

\subsection{Choice of Teacher Model for Distillation}
\label{discuss:teacher}

\begin{figure}[h]
\centering
\includegraphics[width=0.8\linewidth,keepaspectratio]{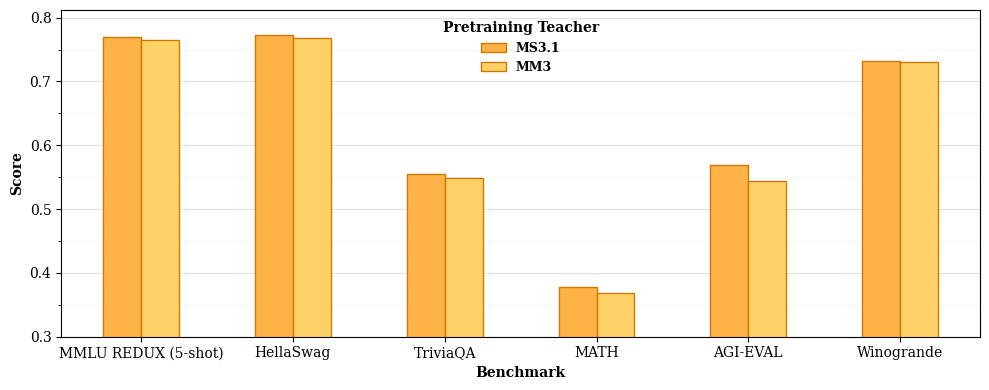}
\caption{\modelname 14B pretraining ablations comparing distillation from Mistral Small 3.1 and Mistral Medium 3 teachers. Despite Mistral Medium 3 being larger and more capable, distillation from Mistral Small 3.1 consistently yields stronger downstream performance across different benchmarks.}
\label{fig:pt_ms3_vs_mm3_teacher}
\end{figure}

In selecting an appropriate teacher model for the distillation process, we identified several noteworthy observations that meaningfully influenced our design choices:

\paragraph{Stronger teacher does not lead to better results:} For pretraining, distilling from Mistral Small 3.1 outperformed distillation from the much stronger Mistral Medium 3 \footnote{\url{https://docs.mistral.ai/models/mistral-medium-3-1-25-08}} even in a non FLOP-matched setup, similar to observations in \cite{busbridge2025distillation} (Figure \ref{fig:pt_ms3_vs_mm3_teacher}).
However, during post-training, \modelname models benefit from distillation from the more capable Mistral Medium 3.1. \nocite{perez2024mmluRedux} \nocite{zellers2019hellaswag} \nocite{joshi2017triviaqa} \nocite{hendrycks2021measuring} \nocite{zhong2023agieval} \nocite{sakaguchi2021winogrande}

\begin{figure}[h]
\centering
\includegraphics[width=0.8\linewidth,keepaspectratio]{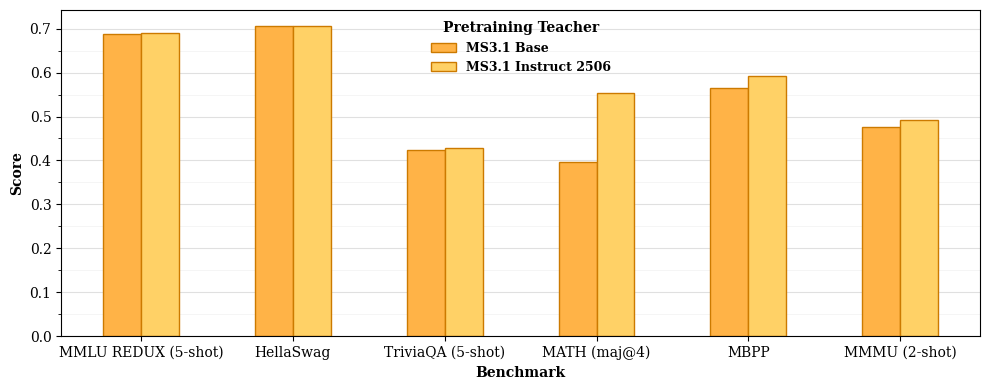}
\caption{\modelname 3B pretraining ablations comparing distillation from base and post-trained (instruct/reasoning) variants of Mistral Small 3.1. The instruct teacher yields stronger performance on STEM benchmarks, while achieving comparable results on knowledge and multimodal evaluations..}
\label{fig:pt_base_vs_instruct_teacher}
\end{figure}

\paragraph{The choice of teacher version (base~/~instruct~) matters:}
\looseness=-1 In line with \citet{goyal2025distilledpretrainingmodernlens}, we find that distilling from a post-trained teacher as opposed to a pre-trained one during the pre-training stage results in a stronger model (Figure \ref{fig:pt_base_vs_instruct_teacher}). In particular, this had a strong impact on maths (MATH) and code capabilities, a small but consistent impact on multimodal evaluations (e.g. MMMU), and a negligible impact on knowledge metrics (MMLU / Trivia-QA). \nocite{perez2024mmluRedux} \nocite{zellers2019hellaswag} \nocite{joshi2017triviaqa} \nocite{hendrycks2021measuring} \nocite{austin2021program} \nocite{yue2023mmmu}

\paragraph{Human Preference tuned models are better teachers:} 
Post 
We use two internal versions of Mistral Medium 3 to answer the question - is it better to distill from an SFT or a preference tuned checkpoint during SFT? We find that distilling from the preference tuned checkpoint is always substantially better. These gains persist even after the student model undergoes its own preference tuning phase.

\subsection{Model Verbosity.}
\label{discuss:verbose}

\begin{figure}[h]
\centering
\includegraphics[width=0.7\linewidth,keepaspectratio]{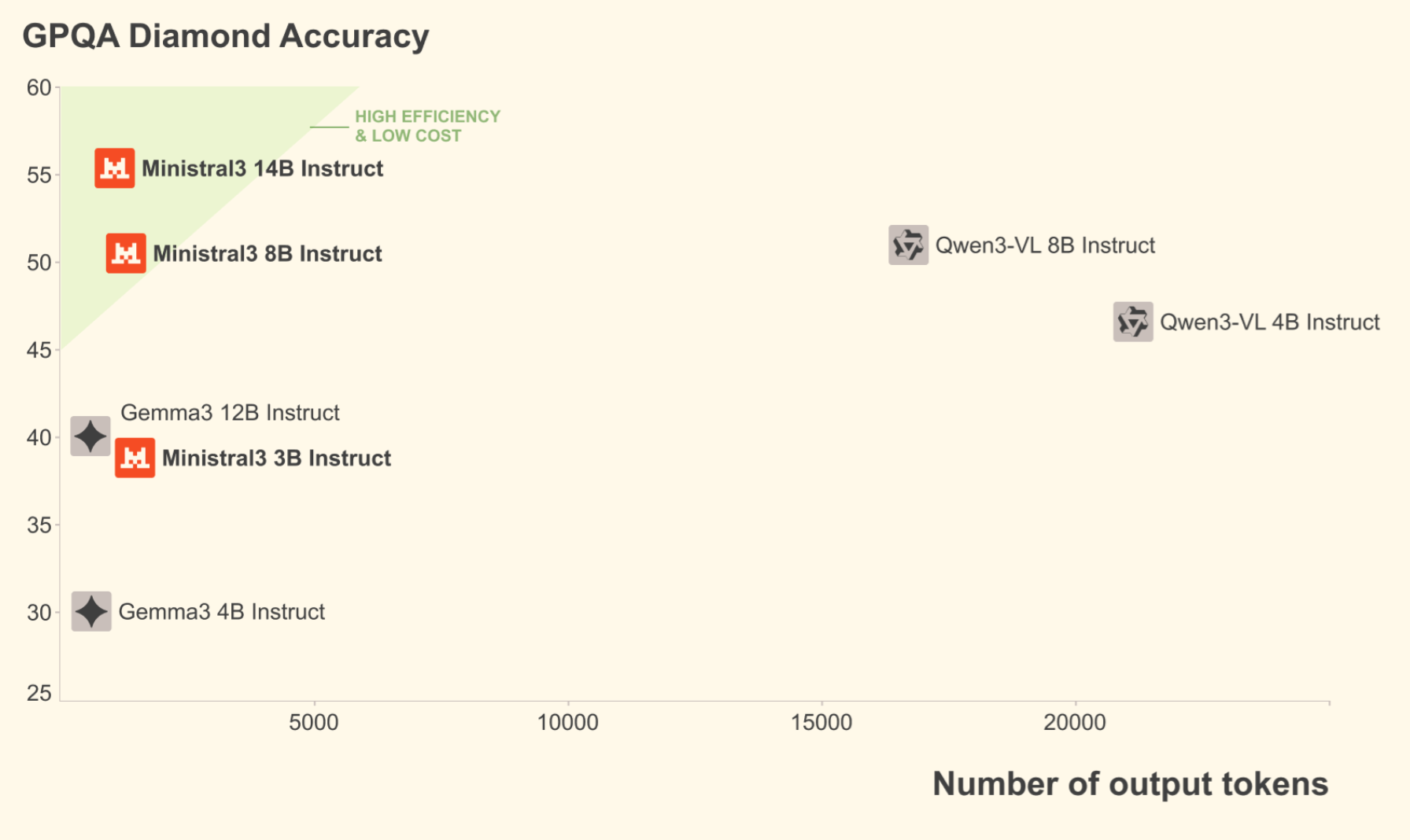}
\caption{Verbosity (in terms of number of output tokens) v.s. accuracy on GPQA Diamond with \modelname instruction-following and reasoning.}
\label{fig:verbose}
\end{figure}

Our post-training of \modelname Instruct differs from Qwen 3 in that it does not do "Reasoning RL" before the "General RL" stage (see Fig. 1 of \cite{yang2025qwen3technicalreport}) this likely results in different model verbosity between the two models as illustrated in Figure~\ref{fig:verbose}.

In an experiment to try and get the \modelname Instruct models to produce longer chains of thought, we investigated incorporating varying proportions of long chain-of-thought (CoT) reasoning traces, paired with carefully curated system prompts, into the SFT training data. Increasing the fraction of such Long CoT data improved the performance on STEM benchmarks; however, it also leads to excessive reflection, internal monologues and backtracking behavior (as shown below), which is undesirable and unnatural for a general-purpose chat model.

\begin{center}
\scalebox{0.85}{%
\begin{tcolorbox}[
    colback=orange!10,
    colframe=orange!80!black,
    arc=4pt,
    boxrule=1pt,
    left=6pt,
    right=6pt,
    top=6pt,
    bottom=6pt,
    fonttitle=\bfseries,
    title=Reasoning Behavior in Instruct Model,
    label={box:unwanted-reasoning}
]

Okay, let's try to solve this problem step by step. The question is: Find the sum of all integer bases \( b > 9 \) for which \( 17_b \) is a divisor of \( 97_b \). First, I need to recall how to convert numbers from base \( b \) to base 10. A number \( d_n d_{n-1} \dots d_1 d_0 \) in base \( b \) is equal to \( d_n b^n + d_{n-1} b^{n-1} + \dots + d_1 b + d_0 \) in base 10. 

\dots

Wait, the problem states \( b > 9 \), so we need to check the divisors of 56 that are greater than 9 (since \( b = d - 7 \)

\dots

Wait, but let's check another way. Instead of using the congruence, we can set up the equation \( 9b + 7 = k(b + 7) \) for some integer \( k \geq 1 \)

\dots

I think that's it. The key was to convert the numbers to base 10, set up the divisibility condition, and recognize that the base must be greater than 9, so we only consider divisors of 56 that make \( b > 9 \). 

\dots

Thus, the valid bases are 21 and 49. The sum of these bases is:
\[ 21 + 49 = 70 \]
Therefore, the answer is \(\boxed{70}\).
\end{tcolorbox}
}
\end{center}

\subsection{ODPO for \modelname Reasoning.}
\label{discuss:odpo}

\begin{figure}[h]
\centering
\includegraphics[width=0.75\linewidth,keepaspectratio]{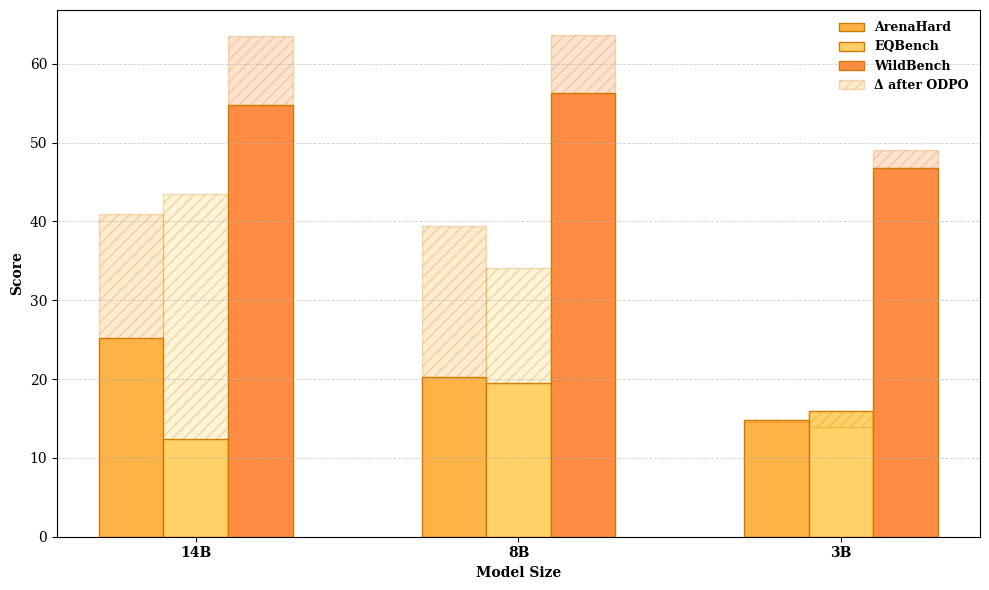}
\caption{Impact of ODPO on chat benchmarks for \modelname reasoning models, applied on top of GRPO-trained checkpoints. ODPO delivers substantial gains across all benchmarks for the 14B and 8B variants.}
\label{fig:reasoning_odpo}
\end{figure}

Reasoning models, while being better at solving challenging problems, often lag in general conversational quality, a pattern we also observed with the \modelname~reasoning variants.
To address this, we performed ODPO training on top of the RL-trained checkpoints.
As shown in Figure~\ref{fig:reasoning_odpo}, this significantly improved the 14B and 8B models on alignment benchmarks.
The 3B model however, did not demonstrate significant improvements on public benchmarks after this stage\footnote{We also found the 3B base more sensitive than 14B and 8B to hyper-parameter choice in fine-tuning}. The model nevertheless
performed better in our internal human evaluations and so we selected the ODPO checkpoint as the release candidate. \nocite{li2024arenahard} \nocite{paech2023eqbench} \nocite{lin2024wildbench}

\section{Conclusion}
\label{sec:conclusion}

We introduced \modelname, a family of efficient dense language models designed for resource-constrained environments. Through iterative distillation from larger teacher models (Mistral Small 3.1 and Medium 3), we created three model sizes (14B, 8B, 3B) each available in base, instruction-following, and reasoning-enhanced variants.
All models support vision capabilities and handle contexts up to 256K tokens.
Collectively, \modelname models highlight Mistral’s continued commitment to supporting and advancing open-source initiatives.
We hope they will provide value to the community and contribute to a stronger, more vibrant open-source ecosystem.

\subsection*{Core contributors}
Alexander H. Liu, Kartik Khandelwal, Sandeep Subramanian, Victor Jouault

\subsection*{Contributors}
Abhinav Rastogi, Adrien Sadé, Alan Jeffares, Albert Jiang, Alexandre Cahill, Alexandre Gavaudan, Alexandre Sablayrolles, Amélie Héliou, Amos You, Andy Ehrenberg, Andy Lo, Anton Eliseev, Antonia Calvi, Avinash Sooriyarachchi, Baptiste Bout, Baptiste Rozière, Baudouin De Monicault, Clémence Lanfranchi, Corentin Barreau, Cyprien Courtot, Daniele Grattarola, Darius Dabert, Diego de las Casas, Elliot Chane-Sane, Faruk Ahmed, Gabrielle Berrada, Gaëtan Ecrepont, Gauthier Guinet, Georgii Novikov, Guillaume Kunsch, Guillaume Lample, Guillaume Martin, Gunshi Gupta, Jan Ludziejewski, Jason Rute, Joachim Studnia, Jonas Amar, Joséphine Delas, Josselin Somerville Roberts, Karmesh Yadav, Khyathi Chandu, Kush Jain, Laurence Aitchison, Laurent Fainsin, Léonard Blier, Lingxiao Zhao, Louis Martin, Lucile Saulnier, Luyu Gao, Maarten Buyl, Margaret Jennings, Marie Pellat, Mark Prins, Mathieu Poirée, Mathilde Guillaumin, Matthieu Dinot, Matthieu Futeral, Maxime Darrin, Maximilian Augustin, Mia Chiquier, Michel Schimpf, Nathan Grinsztajn, Neha Gupta, Nikhil Raghuraman, Olivier Bousquet, Olivier Duchenne, Patricia Wang, Patrick von Platen, Paul Jacob, Paul Wambergue, Paula Kurylowicz, Pavankumar Reddy Muddireddy, Philomène Chagniot, Pierre Stock, Pravesh Agrawal, Quentin Torroba, Romain Sauvestre, Roman Soletskyi, Rupert Menneer, Sagar Vaze, Samuel Barry, Sanchit Gandhi, Siddhant Waghjale, Siddharth Gandhi, Soham Ghosh, Srijan Mishra, Sumukh Aithal, Szymon Antoniak, Teven Le Scao, Théo Cachet, Theo Simon Sorg, Thibaut Lavril, Thiziri Nait Saada, Thomas Chabal, Thomas Foubert, Thomas Robert, Thomas Wang, Tim Lawson, Tom Bewley, Tom Bewley, Tom Edwards, Umar Jamil, Umberto Tomasini, Valeriia Nemychnikova, Van Phung, Vincent Maladière, Virgile Richard, Wassim Bouaziz, Wen-Ding Li, William Marshall, Xinghui Li, Xinyu Yang, Yassine El Ouahidi, Yihan Wang, Yunhao Tang, Zaccharie Ramzi

\newpage

\bibliography{ref}
\vfill
\pagebreak

\appendix

\end{document}